%% file: Skin.tex
\documentclass[10pt,twocolumn,letterpaper]{article}

\usepackage[pagenumbers]{wacv} % To force page numbers, e.g. for an arXiv version

\usepackage{graphicx}
\usepackage{amsmath}
\usepackage{amssymb}
\usepackage{booktabs}
\usepackage{multirow}
\usepackage{tabularx}
\usepackage{bm}
\usepackage{color}
\usepackage{ulem}
\usepackage{cite}

\usepackage{pifont}

\usepackage[pagebackref,breaklinks,colorlinks]{hyperref}

\usepackage[capitalize]{cleveref}
\crefname{section}{Sec.}{Secs.}
\Crefname{section}{Section}{Sections}
\Crefname{table}{Table}{Tables}
\crefname{table}{Tab.}{Tabs.}

\def\wacvPaperID{73} % *** Enter the WACV Paper ID here
\def\confName{WACV}
\def\confYear{2025}

\begin{document}

\title{A Novel Perspective for Multi-modal Multi-label Skin Lesion Classification}

\author{Yuan Zhang$^{1}$\thanks{Corresponding author. Email: yuan.zhang01@adelaide.edu.au.}, Yutong Xie$^{1}$, Hu Wang$^{2}$, Jodie C Avery$^{1}$, M Louise Hull$^{1}$, Gustavo Carneiro$^{3}$\\ \\$^{1}$ University of Adelaide, Australia\\$^{2}$Mohamed bin Zayed University of Artificial Intelligence, United Arab Emirates\\$^{3}$ University of Surrey, UK\\}
% \author{Yuan Zhang\\University of Adelaide\\
% {\tt\small yuan.zhang01@adelaide.edu.au}
% \and
% Yutong Xie\\University of Adelaide, Australia\\
% {\tt\small yutong.xie01@adelaide.edu.au}
% \and
% Hu Wang\\Mohamed bin Zayed University of Artificial Intelligence, United Arab Emirates \\
% {\tt\small Hu.Wang@mbzuai.ac.ae}
% \and
% Jodie C Avery\\University of Adelaide, Australia\\
% {\tt\small jodie.avery@adelaide.edu.au}
% \and
% M Louise Hull\\University of Adelaide, Australia\\
% {\tt\small louise.hull@adelaide.edu.au}
% \and
% Gustavo Carneiro\\University of Surrey, UK\\
% {\tt\small g.carneiro@surrey.ac.uk}
% }
\maketitle

%%%%%%%%% ABSTRACT
\begin{abstract}
The efficacy of deep learning-based Computer-Aided Diagnosis (CAD) methods for skin diseases relies on analyzing multiple data modalities (i.e., clinical+dermoscopic images, and patient metadata) and addressing the challenges of multi-label classification. 
Current approaches tend to rely on limited multi-modal techniques and treat the multi-label problem as a multiple multi-class problem, overlooking issues related to imbalanced learning and multi-label correlation.
This paper introduces the innovative \textbf{Skin} Lesion Classifier, utilizing a \textbf{M}ulti-modal \textbf{M}ulti-label Trans\textbf{Former}-based model (\textbf{SkinM2Former}). 
For multi-modal analysis, we introduce the Tri-Modal Cross-attention Transformer (TMCT) that fuses the three image and metadata modalities at various feature levels of a transformer encoder.
For multi-label classification, we introduce a multi-head attention (MHA) module to learn multi-label correlations, complemented by an optimisation that handles multi-label and imbalanced learning problems.
SkinM2Former achieves a mean average accuracy of 77.27\% and a mean diagnostic accuracy of 77.85\% on the public Derm7pt dataset, outperforming state-of-the-art (SOTA) methods.
\end{abstract}

%%%%%%%%% BODY TEXT
\section{Introduction}
\label{sec:intro}

Skin cancer is the most common type of cancer in many countries, with melanoma accounting for over 80\% of skin cancer deaths~\cite{saginala2021epidemiology}. The 5-year survival rate for patients with early-stage melanoma exceeds 99\%, dropping to 35\% upon distant organ metastasis, emphasizing the importance of early detection and timely treatment~\cite{siegel2024cancer, leiter2020epidemiology}.
% Skin cancer, notably melanoma, is prevalent globally, constituting a significant portion of skin cancer deaths, with a 5-year survival rate exceeding 99\% for early-stage melanoma but dropping to 35\% in cases of distant organ metastasis~\cite{siegel2024cancer, leiter2020epidemiology}. 
Dermatologists employ clinical and dermoscopic images, along with patient metadata, for comprehensive pattern analysis during skin cancer diagnosis~\cite{naik2021cutaneous}. 
Specifically, clinical images, captured using a digital camera, provide a macroscopic view with details such as colour, geometry, and overall appearance~\cite{pizzichetta2004amelanotic}. In contrast, dermoscopic images are obtained through dermoscopy, offering a magnified microscopic view of the lesion that uncovers detailed subsurface structures~\cite{kittler2021evolution}. 
Furthermore, patient metadata, including location, gender, and age, supplements this analysis by providing contextual information~\cite{matthews2017epidemiology}.
To simplify the diagnostic process, the 7-Point Checklist~\cite{argenziano2011seven}, based on seven criteria to derive a diagnostic label (8 classification labels in total), is widely used for the primary diagnosis of skin cancer~\cite{walter2013using}. 

% The widely utilized 7-Point Checklist~\cite{argenziano2011seven} simplifies the diagnostic process, incorporating seven criteria to derive eight diagnostic labels for primary skin cancer diagnosis~\cite{walter2013using}.

The importance of early skin cancer detection has motivated the medical image analysis community to explore deep learning-based Computer-Aided Diagnosis (CAD) tools to assist dermatologists in delivering precise diagnoses. 
However, many skin cancer classification methods, predominantly single-modal (dermoscopic image) multi-class approaches~\cite{hasan2023survey}, are tailored to the limited ISIC benchmark~\cite{gutman2016skin, codella2018skin, codella2019skin, siim-isic-melanoma-classification} and are not equipped to analyze multi-modal data or predict the multi-label 7-Point Checklist. 
This limitation hinders their integration into the clinical workflow described earlier, diminishing the likelihood of acceptance by clinicians.

\begin{figure*}[t!]
\centering
\includegraphics[width=0.94\textwidth]{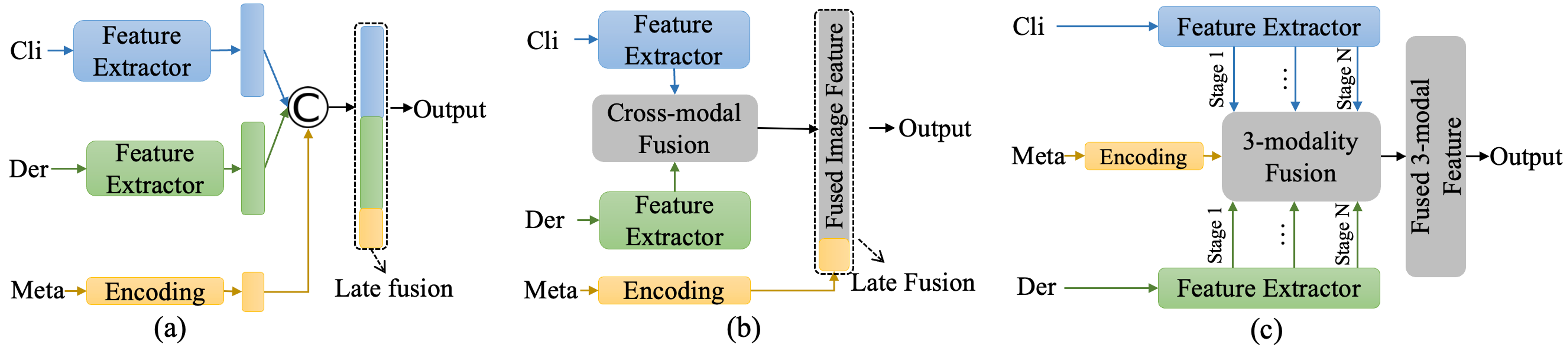}
\caption{Multi-modal skin cancer classifier: (a) late fusion~\cite{kawahara2018seven}; (b) hybrid fusion of image and late fusion of metadata~\cite{tang2022fusionm4net, zhang2023tformer}; (c) our hybrid fusion of all modalities.
%\yuan{I put (a),(b) into the figure itself to save one line space. Is it correct? Do we need to change the font size?}
} \label{fig:fusion_methods}
\end{figure*}
\begin{figure*}[t!]
\centering
\includegraphics[width=0.94\textwidth]{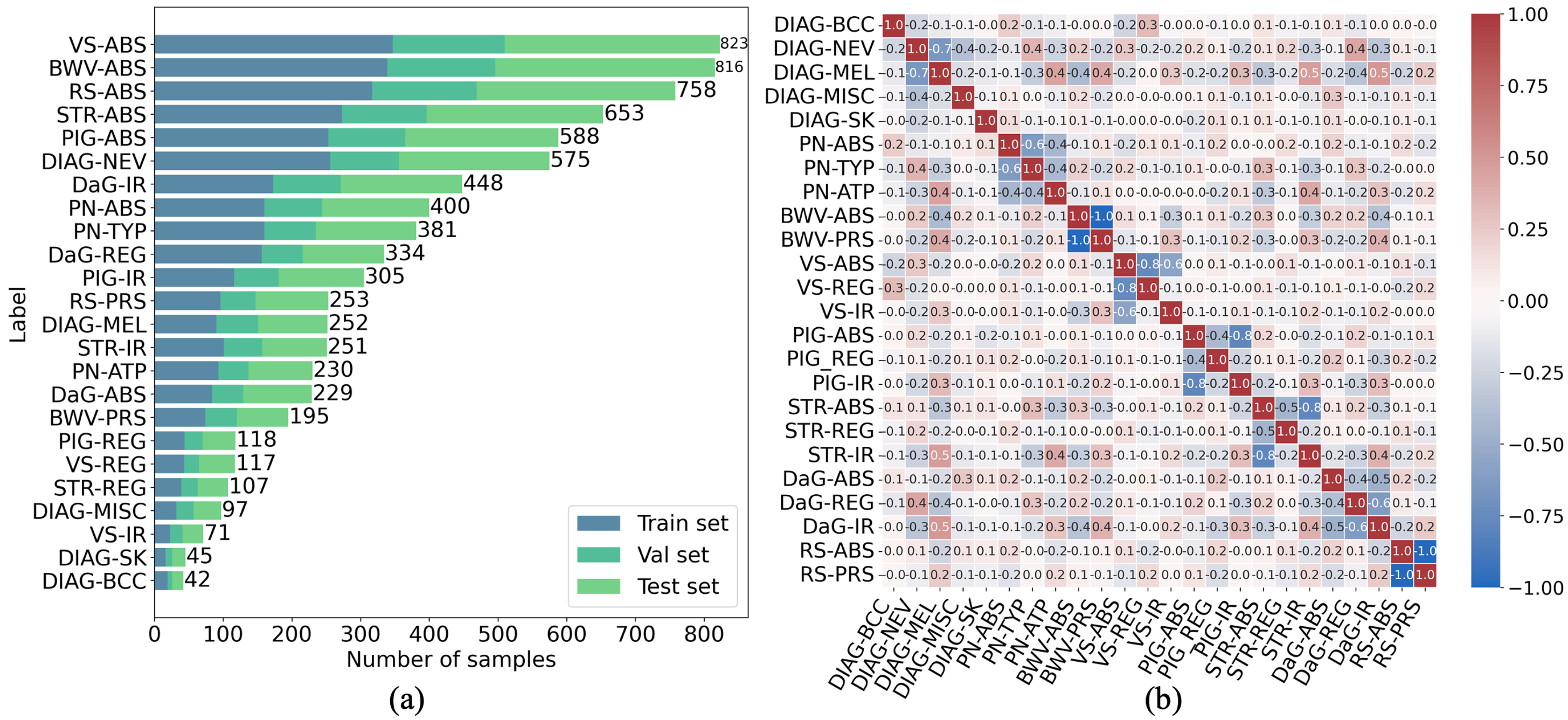}
  \caption{(a) Imbalanced distribution of samples per class. (b) 
  Inter-label Pearson Correlation Coefficients heatmap. Note that labels are denoted by ``Classification problem (\{DIAG,PN,...\})-Possible classes(\{ABS,NEV,...)''. }
  \label{fig:dataset}
\end{figure*}

Limited research exists on the simultaneous fusion of three modalities in skin cancer classification. 
Notable methods include Kawahara et al.'s late fusion using two Inception-V3 networks~\cite{kawahara2018seven} (Figure~\ref{fig:fusion_methods}a), FusionM4Net~\cite{tang2022fusionm4net} and TFormer~\cite{zhang2023tformer} with a hybrid fusion of imaging modalities, followed by a late fusion of metadata (Figure~\ref{fig:fusion_methods}b). 
A limitation of the methods above is that they do not simultaneously fuse the three modalities at multiple feature levels, potentially overlooking important correlations of image and metadata features at distinct hierarchical levels.
Furthermore, existing methods address the multi-label 7-point checklist classification as multiple multi-class learning problems, introducing imbalanced learning and neglecting label correlations~\cite{argenziano2011seven, kawahara2018seven, tang2022fusionm4net, wang2022adversarial, zhang2023tformer}, as Figure~\ref{fig:dataset} shows.

In this paper, we address the automatic skin lesion classification challenges of multi-modality and multi-labeling through the introduction of the \textbf{Skin} Lesion Classifier built upon a \textbf{M}ulti-modal \textbf{M}ulti-label Trans\textbf{Former}-based model (\textbf{SkinM2Former}).  
The multi-modal challenge is addressed with a new Tri-Modal Cross-attention Transformer (TMCT) module to fuse dermoscopic image, clinical image, and metadata at multiple feature levels of a Swin transformer-based architecture~\cite{liu2021swin}, as Figure~\ref{fig:fusion_methods}c shows.
To handle the multi-label classification, we introduce a new attention mechanism to find label associations and a multi-label training~\cite{kobayashi2023two} that is robust to imbalanced learning.
%In the decision-making phase, we utilize an attention mechanism to uncover the associations between labels. Additionally, we employ a loss function designed to expand the classification margin between hard positives and hard negatives from both sample-wise and class-wise ways, achieving efficient multi-label classification performance. 
Our contributions are:
\begin{itemize}
    \item The new TMCT module to fuse multiple feature levels (from low to high level) of the three input data modalities (i.e., clinical image, dermoscopic image, and metadata).
    \item An innovative multi-label model comprising a new attention mechanism that learns the associations between different labels, and a multi-label training strategy that is robust to imbalanced class distributions~\cite{kobayashi2023two}.
    %To boost classification performance from a multi-label classification perspective while remaining high-label-granularity, we employ an attention mechanism to consider the associations between different labels and a loss function to enlarge the classification margin between positive and negative classes from both sample and class perspectives. 
    % To the best of our knowledge, this is the first skin classification study that simultaneously focuses on multimodal fusion and high-label-granularity multi-label classification.
\end{itemize}
Experimental results on the public Derm7pt dataset demonstrate that our SkinM2Former achieves a mean accuracy of 77.27\% and a mean diagnostic accuracy of 77.85\%, which is significantly better than state-of-the-art (SOTA) methods.

\section{Related Works}
\subsection{Automated Diagnosis of Skin Lesion}
Computer-aided methods for classifying skin lesion images have drawn significant research attention because automated analysis can offer patients timely and consistent diagnoses, particularly in remote areas with limited access to clinical services.
Previous research has primarily focused on utilizing a single dermoscopic modality. 
Bayasi et al. \cite{bayasi2023continual} propose Continual-GEN, a subnetwork-based sequential learning method for skin lesion classification that trains with five datasets. 
%\gustavo{instead of focusing only on the Derm7pt dataset, as performed by our paper and the current skin disease SOTA classifiers}.\yuan{Do we need to emphasize it here? As none of these methods in this section used only Derm7pt, my plan was to introduce recent works about single-modal skin classification methods.}
Kanca et al.~\cite{kanca2022learning} develop a K-nearest neighbour algorithm that utilizes manually extracted features from the lesion areas on segmented dermoscopic images to classify melanoma, nevus, and seborrheic keratosis. 
Wang et al.~\cite{wang2023ssd} introduce a dual relational knowledge distillation framework SSD-KD to unify diverse knowledge for skin disease classification.
However, these methods do not fully align with the standard diagnostic practices of dermatologists, which involve combining clinical images, dermoscopic images and patients' metadata to derive a comprehensive diagnosis. To bridge the gap, Derm7pt~\cite{kawahara2018seven} was published as the first large-scale multi-modal and multi-label skin lesion classification dataset, thereby establishing a foundational framework for subsequent research in automated skin lesion classification.

\subsection{Multi-modal Skin Lesion Classification}
Multi-modal skin lesion classification research is becoming more prevalent.
Patrício et al. \cite{patricio2023coherent} present CBE, a concept-based model for skin lesion diagnosis that integrates segmentation-based attention and coherence loss to enhance interpretability and classification accuracy.
Bie et al.\cite{bie2024mica} propose a multi-level image-concept alignment framework MICA for explainable skin lesion classification, which combines a vision model with a language model and aligns the concept information with image semantic characteristics.
However, these methods used only a subset of the Derm7pt dataset (827 images of Nevus and Melanoma), without focusing on the multi-label classification problem.
%However, these methods used only a subset of the Derm7pt dataset (827 images of Nevus and Melanoma), \sout{with an ultimate goal of binary classification} \gustavo{focusing on multiple binary classifications instead of multi-label classification}.\yuan{These two papers solve a binary classification problem (“Nevus” vs “Melanoma”), the "concepts" they used to align with images/segmentations are 7-Point Checklist, but the prediction of those labels is more like an auxiliary task. So it's a bit hard to define if it's multi-label or something else}

For methods focusing on the 7-Point Checklist, most employ a late fusion strategy to jointly analyse complementary information from different modalities.
Kawahara et al.~\cite{kawahara2018seven} propose the use of two Inception-V3 networks to extract clinical and dermoscopic image features that are concatenated with metadata in a late fusion approach. 
Tang et al.~\cite{tang2022fusionm4net} present FusionM4Net, a multi-stage training approach that first extracts decision information using two CNN models from image modalities, then incorporates metadata at a later stage in the model.
Alternatively, Zhang et al.~\cite{zhang2023tformer} employ TFormer with a hybrid fusion strategy, initially fusing the two image modalities at the feature level and subsequently integrating metadata for classification also at a later stage of the model.
However, these approaches do not concurrently integrate all three modalities at multiple feature levels, potentially missing significant correlations between image characteristics and metadata. 

\subsection{Multi-label Skin Lesion Classification with Imbalanced Learning}
There is minimal focus on skin lesion classification from a multi-label perspective.
The multi-label skin lesion classification in most previous studies has been handled by decomposing it into several independent multi-class tasks~\cite{argenziano2011seven, kawahara2018seven, tang2022fusionm4net, wang2022adversarial, zhang2023tformer}, resulting in imbalance issues and ignoring label correlations.
Effectively modelling label correlation and feature-label dependencies can reduce the complexity of learning processes and the dimension of the prediction space~\cite{che2022label}.
Wang et al.~\cite{wang2021incorporating} utilise constrained classifier chain (CC) to enhance multi-label classification performance by minimizing binary cross-entropy (BCE) loss.
However, CC requires each classifier to be binary, leading to the reassignment of originally multiple multi-class labels into a bunch of binary multi-labels, which compromises label precision.

Furthermore, the BCE loss introduces significant class imbalance issues. Mitigating these issues can prevent the model from overly emphasizing categories with more samples during training, thereby enhancing its generalization ability in real clinical scenarios~\cite{tarekegn2021review, kobayashi2023two}. 
Previous research tackled the imbalance issue associated with BCE by employing simple methods such as frequency-based weighting~\cite{pham2021learning} and adaptive weighting schemes~\cite{ridnik2021asymmetric, lin2017focal}. 
Therefore, exploring label correlations and utilizing advanced loss functions that bypass imbalance issues while enhancing discriminativity could serve as fundamental steps in studying skin classification from a multi-label perspective.

\section{Methodology}

\begin{figure*}[t!]
\centering
\includegraphics[width=0.90\textwidth]{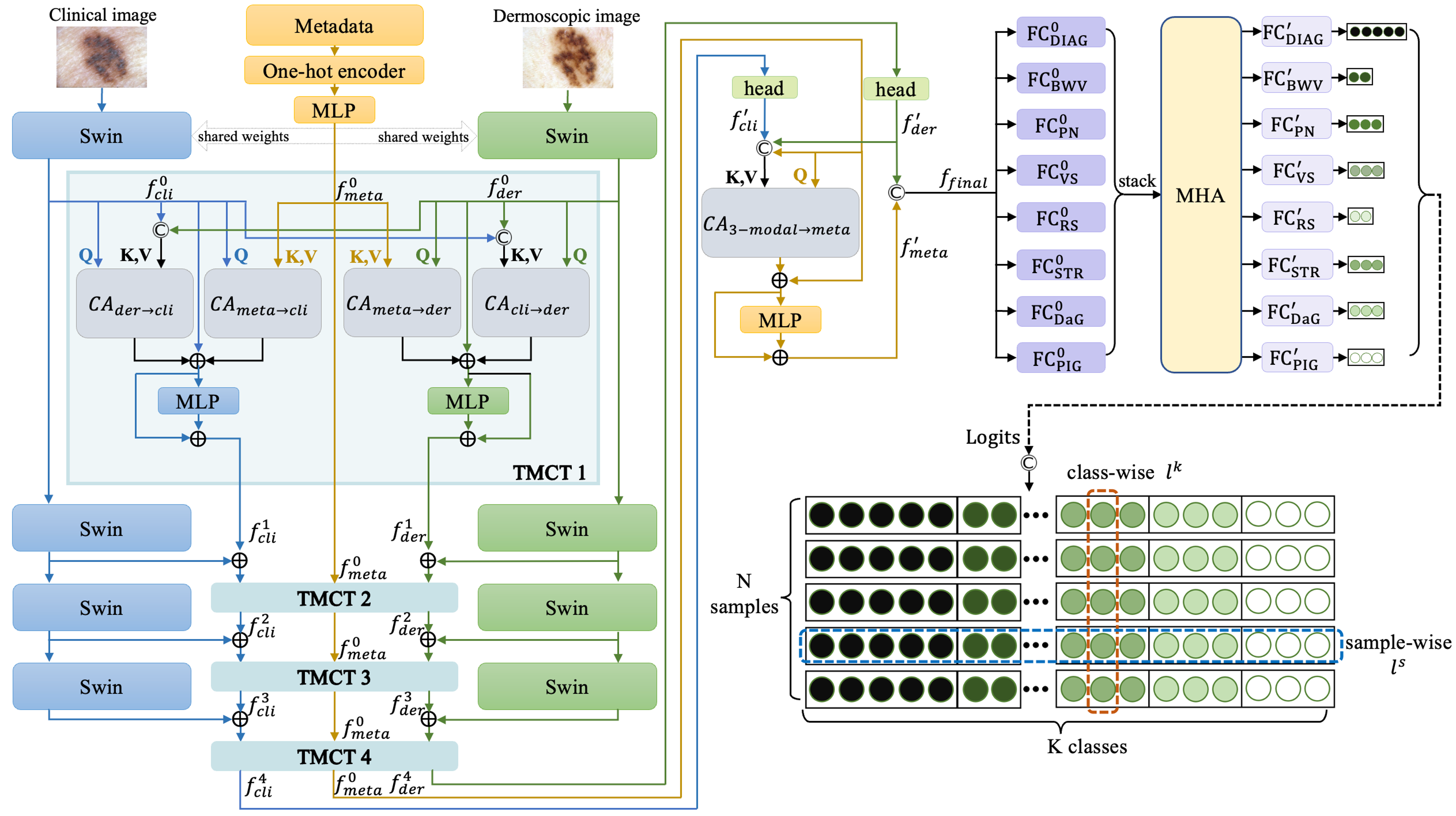}
\caption{\textbf{SkinM2Former}: Tri-modal Cross-attention Transformer (TMCT) module to fuse all modalities; multi-head attention (MHA) layer to learn multi-label associations; and multi-label loss~\cite{kobayashi2023two} robust to class imbalances.} \label{fig1}
\end{figure*}

The proposed SkinM2Former (Figure~\ref{fig1}) has three components: a Tri-Modal Cross-attention Transformer (TMCT) to integrate the information from clinical images, dermoscopic images, and metadata at multiple feature levels; a Multi-Head Attention (MHA) module to capture correlations between labels; and a multi-label training method that is robust to class-distribution imbalances.

Let us represent the multi-modal skin-cancer dataset as $\mathcal{D}=\{ (\mathbf{x}^{cli}_i, \mathbf{x}^{der}_i, \mathbf{x}^{meta}_i, \allowbreak \mathbf{y}_i) \}_{i=1}^{N}$ (e.g., Derm7t dataset~\cite{kawahara2018seven}), with $N$ clinical images $\mathbf{x}^{cli} \in \mathcal{X}^{cli} \subset  \mathbb{R}^{W \times H \times C}$ ($W$, $H$ and $C$ are height, width and channels), dermoscopic images $\mathbf{x}^{der} \in \mathcal{X}^{der} \subset  \mathbb{R}^{W \times H \times C}$ and metadata $\mathbf{x}^{meta} \in \mathcal{X}^{meta} \subset  \mathbb{R}^{L}$ ($L$ is the amount of metadata covariates), annotated with a set of multi-class labels $\mathbf{y}_i=\{ (\mathbf{y}^{DIAG}_i, \mathbf{y}^{BWV}_i,  \allowbreak  \mathbf{y}^{PN}_i, \mathbf{y}^{VS}_i, \mathbf{y}^{RS}_i, \mathbf{y}^{STR}_i,\mathbf{y}^{DaG}_i, \mathbf{y}^{PIG}_i)\}_{i=1}^{N}$.
The labels in $\mathbf{y}_i$ represent one diagnosis label $\mathbf{y}^{DIAG} \in  \{0,1\}^5$, and seven skin lesion attribute labels for each sample, where $\mathbf{y}^{BWV} \in  \{0,1\}^2$, $\mathbf{y}^{PN} \in  \{0,1\}^3$, $\mathbf{y}^{VS} \in  \{0,1\}^3$, $\mathbf{y}^{RS} \in  \{0,1\}^2$, $\mathbf{y}^{STR} \in  \{0,1\}^3$, $\mathbf{y}^{DaG} \in  \{0,1\}^3$, and $\mathbf{y}^{PIG} \in  \{0,1\}^3$.

\subsection{Tri-Modal Cross-modality Fusion}

The proposed TMCT module (Figure~\ref{fig1}) integrates multi-scale clinical and dermoscopic image features (leveraging Swin Transformer\cite{liu2021swin} for its effective long-range dependency capture~\cite{han2022survey}) with metadata features from a multi-layer perceptron (MLP). In the first TMCT module, clinical image features $f_{cli}^0$ and dermoscopic image features $f_{der}^0$ from the first Swin Transformer block, alongside metadata feature $f_{meta}^0$, are fused using cross-modal attention (CA) blocks. The details of ${CA}_{der \rightarrow  cli}$ \& ${CA}_{meta \rightarrow cli}$ and ${CA}_{cli \rightarrow der }$ \& ${CA}_{meta \rightarrow der}$ are the same, so we only describe ${CA}_{der \rightarrow cli}$, ${CA}_{meta \rightarrow cli}$. 
The ${CA}_{der \rightarrow cli}$ uses $f_{cli}^0$ as the query and the concatenation of $f_{cli}^0$ and $f_{der}^0$ (denoted by $[f_{cli}^0, f_{der}^0]$) as the key and value~\cite{zhang2023tformer}; and
for ${CA}_{meta \rightarrow cli}$, we apply a multi-head attention layer which takes $f_{cli}^0$ as the query and $f_{meta}^0$ as the key and value, calculated as
\begin{equation}
\scalebox{0.9}{$
\begin{split}
CA_{der \rightarrow cli} &= \mathsf{window\_multihead}(f_{cli}^0, f_{der}^0),
\end{split}
$}
\end{equation}
\begin{equation}
\scalebox{0.9}{$
\begin{split}
CA_{meta \rightarrow cli} = \mathsf{\mathsf{multihead}}(f_{cli}^0, f_{meta}^0).
\end{split}
$}
\end{equation}

The output of ${CA}_{der \rightarrow cli}$ and ${CA}_{meta \rightarrow cli}$ are combined and connected with the initial input $f_{cli}^0$ via a residual operation, then passed to an MLP with a residual shortcut. Finally, we have the new clinical image future $f_{cli}^1$ which integrates the information from both dermoscopic image and metadata as follows:
\begin{equation}
\scalebox{0.9}{$
\begin{aligned}
f_{cli}^1 = & \ \mathsf{MLP}\left( CA_{der \rightarrow cli} + CA_{meta \rightarrow cli} + f_{cli}^0 \right) \\
            & + CA_{der \rightarrow cli} + CA_{meta \rightarrow cli} + f_{cli}^0.
\end{aligned}
$}
\end{equation}

%\gustavo{Did we address the issues in Q5: We apologize for the confusion. we will clarify the abbreviation TWL as "Two Way Loss” and correct the typos in Eq.2 and Eq.3 as accurately depicted in Fig.3. As for point 3 of R4, our description in Fig.3 and Eq.4 is the general setting, but the text emphasises the optimal setting. We will update the figure and equation to reflect the optimal setting in the final version.}\yuan{Yes, I have updated all of them}

The internal representations of the clinical image are refined through dual cross-attention blocks, effectively incorporating features from dermoscopic images and metadata. This synergistic fusion within a latent space enhances the model's adaptability and utilization of complementary information across modalities. The same fusion strategy is applied to dermoscopic images to yield $f_{der}^1$. The outputs of the first TMCT module are $f_{cli}^1$, $f_{der}^1$, and a copy of $f_{meta}^0$.
To improve the capability of capturing cross-modal information across varying scales, we combine the features extracted by each modality backbone to $f_{cli}^1$ and $f_{der}^1$ respectively, then compound with $f_{meta}^0$, as the input of the next TMCT module. We apply one TMCT module after each stage of the backbone, to gradually fuse the features from low to high level. After the fourth TMCT module, the output $f_{cli}^4$ and $f_{der}^4$ are fed into two head modules consisting of a pooling layer and a fully connected layer to get the final fused image feature vectors \(f_{cli}^{\prime}\) and \(f_{der}^{\prime}\).

To enhance further the feature fusion between image and non-image modalities, we apply an attention-based transformer block ${CA}_{3-modal \rightarrow meta}$ which takes meta-data feature $f_{meta}^0$ as query and the concatenation of \(f_{cli}^{\prime}\), \(f_{der}^{\prime},f_{meta}^0\) as key and value, with ${CA}_{3-modal \rightarrow meta} = \mathsf{multihead}\left(f_{meta}^0, [f_{cli}^{\prime}, f_{der}^{\prime}, f_{meta}^0]\right)$.

Then, we feed the output into an MLP with a residual shortcut to get the fused metadata feature \( f_{meta}^{\prime} \) as follows:
\begin{equation}
\scalebox{0.9}{$
\begin{aligned}
f_{meta}' = \mathsf{MLP} \left( \mathsf{multihead}\left(f_{meta}^0, [f_{cli}', f_{der}', f_{meta}^0]\right) + f_{meta}^0 \right) \\  + \mathsf{multihead}\left(f_{meta}^0, [f_{cli}', f_{der}', f_{meta}^0]\right)+ f_{meta}^0.
\end{aligned}
$}
\end{equation}

After this Tri-Modal cross-modality fusion, we get the feature \( f_{final} \) for the following classification module by concatenating dermoscopic feature \( f_{der}^{\prime} \) and the metadata feature \( f_{meta}^{\prime} \) as follows:
\begin{align}
f_{final} &= [f_{der}^{\prime}, f_{meta}^{\prime}].
\end{align}

\subsection{Multi-label Classification}
We employ a Multi-Head Attention (MHA) layer to estimate the correlations among various attributes. Specifically, we pass the classification feature \( f_{final} \) into a set of 8 fully connected (FC) layers to get the initial classification features for each attribute. These features are stacked and passed to the MHA. The output from the MHA is subsequently unstacked and input into another set of 8 classification layers to yield the multi-label probabilities $\mathbf{p}_i=\{ (\mathbf{p}^{DIAG}_i, \mathbf{p}^{PN}_i, \mathbf{p}^{BWV}_i, \allowbreak \mathbf{p}^{VS}_i, \mathbf{p}^{PIG}_i, \mathbf{p}^{STR}_i,  \allowbreak \mathbf{p}^{DaG}_i, \mathbf{p}^{RS}_i)\}_{i=1}^{N}$ for the diagnosis of skin diseases. 
%\yuan{Could we perhaps elaborate more on why TWL is effective for addressing class imbalance issues? I'm struggling to come up with the right words to explain it clearly and would really appreciate your help.}
Assuming that the logit vectors to produce these probabilities are represented by
$\mathbf{l}_i=\{ (\mathbf{l}^{DIAG}_i, \mathbf{l}^{PN}_i, \mathbf{l}^{BWV}_i, \mathbf{l}^{VS}_i, \allowbreak \mathbf{l}^{PIG}_i, \mathbf{l}^{STR}_i, \mathbf{l}^{DaG}_i, \mathbf{l}^{RS}_i)\}_{i=1}^{N}$, we train our model with the Two-Way Loss function (TWL), that is shown to be robust to the class imbalances present in multi-label learning~\cite{kobayashi2023two}:
\begin{equation}
\scalebox{0.85}{$
\begin{aligned}
\ell = \frac{1}{N} \sum_{i=1}^{N} \ell_{sp}(\{(\mathbf{l}_{ik}, \mathbf{y}_{ik})\}_{k=1}^{K}; T) + \frac{1}{K} \sum_{k=1}^{K} \ell_{sp}(\{(\mathbf{l}_{ik}, \mathbf{y}_{ik})\}_{i=1}^{N}; T),
\end{aligned}
$}
\end{equation}
which is optimised over the $N$ samples and $K$ labels using 
\begin{equation}
\scalebox{0.9}{$
\begin{aligned}
\ell_{sp}(\{(\mathbf{l}_{ik}, \mathbf{y}_{ik})\}_{k=1}^{K}; T) &= \\
\mathsf{softplus} \left( \log \sum_{n=1|y_{in}=0}^{K} e^{\mathbf{l}_{in}} \right. 
&\quad + \left. T \log \sum_{p=1|y_{ip}=1}^{K} e^{-\frac{\mathbf{l}_{ip}}{T}} \right),  
\end{aligned}
$}
\end{equation}
\begin{equation}
\scalebox{0.9}{$
\begin{aligned}
 \text{and } \ell_{sp}(\{(\mathbf{l}_{ik}, \mathbf{y}_{ik})\}_{i=1}^{N}; T) &= \\
\mathsf{softplus} \left( \log \sum_{i=1 | \mathbf{y}_{ik}=0}^{N} e^{\mathbf{l}_{ik}} \right.
&\quad + \left. T \log \sum_{j=1|\mathbf{y}_{jk}=1}^{N} e^{-\frac{\mathbf{l}_{jc}}{T}} \right).
\end{aligned}
$}
\end{equation}
This TWL function addresses class imbalance in multi-label learning by discriminating both classes and samples, which mitigates the imbalance problem more effectively than traditional BCE loss~\cite{kobayashi2023two}.

During the evaluation, we compute $\mathbf{p}$, where the label for each of the 8 attributes is computed as the one with maximum probability.

\section{Experiments}
\subsection{Dataset and Implementation Details}
We evaluate our SkinM2Former on the Derm7pt dataset~\cite{kawahara2018seven},
which is the main medical dataset for the multi-modal, multi-label, imbalanced learning task.
It containing 1011 cases, each comprising a dermoscopic image, a clinical image, patient's metadata, and 8 labels (7-point checklist labels and a diagnosis label). The 7-point checklist labels include: blue whitish veil (BWV), pigment network (PN), vascular structures (VS), regression structures (RS), streaks (STR), dots and globules (DaG), and pigmentation (PIG). Each label has different types, including: Present (PRE), Absent (ASB), Typical (TYP), Atypical (ATP), Regular (REG), and Irregular (IR). The diagnosis (DIAG) label is divided into five types:  Basal Cell Carcinoma (BCC), Nevus (NEV), Melanoma (MEL), Miscellaneous (MISC), and Seborrheic Keratosis (SK). 
More details of the dataset are in~\cite{kawahara2018seven}.  These 1011 cases are split into 413 training, 203 validation and 395 testing cases by~\cite{kawahara2018seven}. Table~\ref{tab:statistics} lists details of dataset distribution.
\input{tables/table1}

Our model was trained for 200 epochs with a batch size of 32, using Adam optimizer~\cite{kingma2014adam} with an initial learning rate of 1e-4 and a weight decay of 1e-4. 
The CosineAnnealingLR schedule is applied for the learning rate decay. 
The model with the best average accuracy on the validation set is saved for testing. 
The ImageNet-1K~\cite{russakovsky2015imagenet} pre-trained Swin-Tiny is employed as the backbone. 
The dimension of both heads, all MLPs, and all FC layers is 128. 
The number of heads of MHA is 4. 
The temperature parameter $T=4$. Input images are resized to 224 × 224 × 3, and the length of the encoded patient's meta-data is 20. Data augmentation consists of random vertical and horizontal flips, shifts, distortions and mixup~\cite{engelmann2022detecting}. All experiments were performed using Python 3.9 with PyTorch 1.12.1 and run on 4 NVIDIA RTX 3090 GPUs with 24 GB VRAM.

Following~\cite{kawahara2018seven, bi2020multi}, our approach is evaluated with the average accuracy (AVG) of all classes. 
We also show F1-score results in Table~2 of the supplementary material since this is a valuable comparison measure, but not as common as AVG.  
All experiments are run 10 times, from which we report the mean and standard deviation values.

\subsection{Experimental Results}
\subsubsection{Comparisons with SOTA}
\input{tables/table2}
\input{tables/table3}

We compare the performance of our method to the following SOTA methods: Inception-unbalanced (Inception\_UB), Inception-balanced (Inception\_B) and Inception-combine (Inception\_CB) methods proposed by~\cite{kawahara2018seven}, MmA~\cite{ngiam2011multimodal}, TripleNet~\cite{ge2017skin}, EmbeddingNet~\cite{yap2018multimodal}, HcCNN~\cite{bi2020multi}, AMFAM~\cite{wang2022adversarial}, FusionM4Net~\cite{tang2022fusionm4net}, and TFormer~\cite{zhang2023tformer}. 

The results in Table~\ref{tab:comparison} show that SkinM2Former achieves an average accuracy of 77.27\%, representing a statistically significant improvement over the current leading methods, FusionM4Net and TFormer. 
The most competitive method in terms of average accuracy is the FusionM4Net, which requires multi-stage training and searching for decision weights for each classifier using the validation set. Such complex training is a clear disadvantage, compared to our proposed end-to-end training method.
% \begin{figure*}[t!]
% \centering
% \includegraphics[width=0.97\textwidth]{examples6.png}
% \caption{Examples of clinical image, dermoscopic image, and patient metadata, along with the predicted labels from FusionM4Net [20], TFormer [25], and our SkinM2Former. Ground truth (GT) is in the second column of the tables.} \label{fig:examples}
% \end{figure*}

\begin{table*}[h!]
\caption{Examples of clinical image, dermoscopic image, and patient metadata, along with the predicted labels from FusionM4Net~\cite{tang2022fusionm4net}, TFormer~\cite{zhang2023tformer}, and our SkinM2Former. Ground truth (GT) is in the second column of the tables.} 
    \centering
    % First row of tables
    \begin{minipage}{0.32\textwidth}
        \input{sample_images/table_image1}
    \end{minipage}\hfill
    \begin{minipage}{0.33\textwidth}
        \input{sample_images/table_image2}
    \end{minipage}\hfill
    \begin{minipage}{0.33\textwidth}
        \input{sample_images/table_image3}
    \end{minipage}

    \vspace{0.5em} % Adjust vertical space between rows as needed

    % Second row of tables
    \begin{minipage}{0.32\textwidth}
        \input{sample_images/table_image4}
    \end{minipage}\hfill
    \begin{minipage}{0.33\textwidth}
        \input{sample_images/table_image5}
    \end{minipage}\hfill
    \begin{minipage}{0.34\textwidth}
        \input{sample_images/table_image6}
    \end{minipage}
    \label{tab:examples}
\end{table*}

%\yuan{New}
Moreover, the results in Table~\ref{tab:F1} indicate that our method also achieved the highest F1-score on the majority of labels, which further displays the effectiveness of our method.

%\yuan{New}
Table~\ref{tab:examples} illustrates the clinical images, dermoscopic images, and detailed patient metadata of 6 examples, along with their predicted labels generated by FusionM4Net [20], TFormer [25], and our proposed SkinM2Former. Comparing the predicted labels with the most competitive methods,  FusionM4Net~\cite{tang2022fusionm4net} and TFormer~\cite{zhang2023tformer}, reveals that our approach accurately predicts the majority of labels. 

\subsubsection{Ablation study}

\input{tables/table4}
To validate the efficacy of each component in SkinM2Former, we conducted the ablation study in Table~\ref{tab:ablation}. 
%\sout{For the baseline model, we removed the TMCT and MHA in SkinM2Former, and used concatenation to fuse three modality features and replaced the two-way loss (TWL) function as Binary Cross-Entropy (BCE).} 
Our baseline model uses a shared weights (SW) Swin-Tiny (Swin) as the feature extractor for imaging modalities, employing concatenation as the feature fusion strategy and Binary Cross-Entropy (BCE) as the loss function. 
Note that the TMCT module enhances feature extraction and fusion across the three modalities, where the average accuracy increases from 74.92\% to 75.92\%. For the multi-label classification strategies (MHA and TWL), we notice that TMTC+TWL, in comparison with TMTC+BCE, shows a small, but consistent gain of 0.22\%.
The addition of MHA forms TMTC+MHA+TWL (SkinM2Former), which when compared to TMTC+MHA+BCE, shows a substantial improvement of 1.24\%.
Furthermore, TMTC+MHA+TWL, in comparison with TMTC+TWL, shows a remarkable improvement of 1.13\%.
It is worth noticing that TMTC+MHA+TWL improved the average accuracy of the baseline from 74.92\% to 77.27\%.
% \gustavo{Yuan, in all these results, we need to say if we mean the with or w/o shared weights...}\yuan{I made a new table(Tab. 6) with another column to show if the method is shared-weights. Will it be better than the old one(Tab. 5)? I also added this information for the baseline model. Our framework figure showed clearly that TMCT is share weights, so I guess we probably don't need texts for TMCT? }\gustavo{Looks good.  Can you then erase Table 5 and make sure the text is referring to the correct new Table 6?}
%did not directly enhance the classification capability of the model, either for the baseline model or for SkinM2Former. 
%However, the combination of both significantly increased the average accuracy, boosting it by 1.12\% and 1.35\% for the baseline model and SkinM2Former, respectively. 
These results support our claims about the TMCT (fusion of multiple feature levels), MHA (interrelations among labels) and TWL (multi-label learning).
%method that leverages the interrelations among labels (MHA) and addresses multi-label classification with the two-way loss (TWL)~\cite{kobayashi2023two}.

Moreover, the results from the third-to-last row of Table~\ref{tab:ablation} indicate that, within the TMCT module, utilizing $f_{meta}^0$ directly as key and value in ${CA}_{meta \rightarrow cli}$ and ${CA}_{meta \rightarrow der}$ leads to higher classification accuracy compared to using the concatenation of $f_{meta}^0$ and image modalities. Therefore, we employ $f_{meta}^0$ as the key and value for ${CA}_{meta \rightarrow cli}$ and ${CA}_{meta \rightarrow der}$ in all experiments.
Additionally, comparing the results between the last and the second-to-last rows of Table~\ref{tab:ablation}, we see that using independent feature extractors for two image modalities led to a 0.83\% decrease in average accuracy, demonstrating the effectiveness of Swin Transformers with shared weights across different modalities. 
Conclusively, using shared weights in feature extractors not only reduces the number of parameters but also enhances the ability to capture and integrate features from both dermoscopic and clinical images.
% \gustavo{Yuan, please check Table~\ref{tab:ablation} since I changed the order of the last three results.}
% \gustavo{Yuan, can you adapt the comments here to refer to new Table 6 instead of Table 5?}

%\gustavo{Are we showing the results for shared weights (Q4 from rebuttal): We experimented with using independent feature extractors for two image modalities, which showed a worse result (mean 76.44/std 0.68) compared to using shared weights (mean 77.27/std 0.47). Unfortunately, the comparisons were omitted due to page limits.}\yuan{Yes, I have added it in the last sentence of the last paragraph.}

\subsubsection{Analysis for multi-modality feature fusion}
\input{tables/table5}
%\gustavo{Moved this experiment to ablation since I think it's more about testing the multi-modal component of the method} 
We compare the use of unimodal and multi-modal data in Table~\ref{tab:single_multi}. Unimodal results for each modality show relatively low accuracy across different labels, with dermoscopic images outperforming metadata, which surpasses clinical images. %Dermoscopic images generally outperform the other two modalities for most labels, but clinical images show the best performance in classifying VS. 
The multi-modal data results suggest that the fusion of two or three modalities enhances accuracy, revealing that diverse modalities contain complementary information that are advantageous for classification.
%and confirms the effectiveness of SkinM2Former in integrating information encapsulated within various modalities. 
In fact, our proposed fusion of the three modalities at multiple feature levels produces the best classification accuracy of 77.27\%.

\subsubsection{Analysis for multi-modality decision fusion}
\input{tables/table6}
Table~\ref{tab:final_f} shows a comparison of different concatenations of decision features from distinct modalities (\( f_{cli}' \), \( f_{der}' \) and \( f_{meta}' \)) to build the final decision feature \( f_{final} \).
%Each of these three features already contains information from three modalities and the results 
Results show that SkinM2Former reaches the highest accuracy of 77.27\% when concatenating \( f_{der}' \) and \( f_{meta}' \) to form \( f_{final} \).
Even though the best DIAG accuracy of 78.25\% is achieved by concatenating \( f_{cli}' \),  \( f_{der}' \) and \( f_{meta}' \), such concatenation of the three decision features only reaches an average of 76.82\%. 
This can be explained by the presence of excessive noise in clinical images. 
However, this does not imply that clinical images are dispensable because their contribution may have been implicitly passed to the model through the TMCT module.
%, resulting in some redundancy compared to dermoscopic images. 
% Therefore, we need to make a trade-off depending on the specific circumstances. 
% The concatenation of three decision features presents a more viable option for clinical implementation. 
%In the realm of machine learning, since the main evaluation metric in previous papers is average accuracy, so we used the concatenation of \( f_{cli}' \)  and  \( f_{der}' \) as \( f_{final} \) as for the comparison of other methods.
Given these results, our SkinM2Former relies on the concatenation of \( f_{cli}' \)  and  \( f_{der}' \) to build the \( f_{final} \).

\subsection{Analysis of Model Applicability in Real-world Scenarios}

Our proposed method is highly applicable in real-world clinical settings.
SkinM2Former has been carefully designed, with each component tailored to address challenges associated with multi-modal, multi-label tasks, and class imbalance issues. 
Table~\ref{tab:ablation} demonstrated the effectiveness and importance of these components. Moreover, training our model requires approximately 1 hour, a process that occurs offline. 
Once trained, the model classifies each sample in less than 0.03 seconds.
This speed suggests that our approach is well-suited for integration into routine clinical workflows. 

% \gustavo{should we talk about training and testing time?  Q2 from rebuttal: Our model is carefully designed, with each component specifically tailored to address challenges associated with multi-modal, multi-label tasks, and class imbalance issues. The effectiveness and importance of these components are shown in Tab.2. Moreover, training our model requires approximately 1 hour, a process which occurs offline. Once trained, the model classifies each sample in less than 0.03 seconds. This speed suggests that our approach is well-suited for integration into routine clinical workflows. As for the generalizability, we show enough evidence of the generalization of our method considering the limited availability of datasets for the multimodal, multi-label, imbalanced learning task. As new datasets for the same task in other diseases become available in the future, we will further test the effectiveness of our method, as we discuss in the conclusion.}

\section{Limitations}
Our approach did not consider the presence of potential confounding variables on the images, such as grid scales and hair, which may have an impact on model performance. 
%preprocess the data, but many clinical and dermatoscopic images contain noise such as grid scales and hair, potentially impacting model performance. 
Future research could leverage methods to detect and remove confounding factors~\cite{yan2023towards}.
%to segment lesions and then crop images for classification, or employ image-level weakly supervised segmentation as an auxiliary task during classification to enhance lesion boundary feature perception, further improving classification outcomes. 
Furthermore, a hierarchical relationship exists between 8 main labels and 24 specific labels, which our research did not specifically address. Future studies could utilize Graph Convolutional Networks (GCN) to better capture this hierarchical relationship, thereby enhancing the model's classification capabilities. 

\section{Conclusion}
In this work, we introduced SkinM2Former for multi-modal multi-label classification of skin lesions. We designed a set of TMCT modules to progressively fuse cross-modal information between three modalities at the feature level. Moreover, we formulated the classification problem of skin lesions from a multi-label perspective, proposing to enhance classification performance by exploring label dependencies and employing more effective loss functions. Experimental results on the publicly available Derm7pt multi-modal multi-label dataset demonstrate that our approach outperforms other SOTA methods.
In addition, we hope our research offers insights for the automated detection of other diseases requiring multi-modal multi-label diagnostics. For instance, multifocal diseases like endometriosis require diagnoses of multiple lesions from various medical image modalities (e.g., MRI and ultrasound). 

%%%%%%%%% REFERENCES
{\small
\bibliographystyle{ieee_fullname}
\bibliography{Skin}
}

\end{document}

%% file: tables/table1.tex
\begin{table}[t!]
\centering
\caption{The detailed statistics for the Derm7pt dataset.}
\resizebox{\columnwidth}{!}{ 
\begin{tabular}{@{}lllllll@{}}
\toprule
Label                                       & Name                 & Abbrev & Train & Val & Test & Total \\ \midrule
\multirow{5}{*}{DIAG}           & Basal Cell Carcinoma & BCC    & 19    & 7        & 16   & 42    \\
                                            & Nevus                & NEV    & 256   & 100      & 219  & 575   \\
                                            & Melanoma             & MEL    & 90    & 61       & 101  & 252   \\
                                            & Miscellaneous        & MISC   & 32    & 25       & 40   & 97    \\
                                            & Seborrheic Keratosis & SK     & 16    & 10       & 19   & 45    \\
                                             \midrule
\multirow{2}{*}{BWV}    & Absent               & ABS    & 339   & 157      & 320  & 816   \\
                                            & Present              & PRS    & 74    & 46       & 75   & 195   \\ \midrule
\multirow{3}{*}{PN}       & Absent               & ABS    & 160   & 84       & 156  & 400   \\
                                            & Typical              & TYP    & 160   & 75       & 146  & 381   \\
                                            & Atypical             & ATP    & 93    & 44       & 93   & 230   \\ \midrule
\multirow{3}{*}{VS}   & Absent               & ABS    & 347   & 163      & 313  & 823   \\
                                            & Regular              & REG    & 43    & 22       & 52   & 117   \\
                                            & Irregular            & IR     & 23    & 18       & 39   & 71    \\ \midrule
\multirow{2}{*}{RS} & Absent               & ABS    & 317   & 152      & 289  & 758   \\
                                            & Present              & PRS    & 96    & 51       & 106  & 253   \\ \midrule
\multirow{3}{*}{STR}              & Absent               & ABS    & 273   & 123      & 257  & 653   \\
                                            & Regular              & REG    & 39    & 24       & 44   & 107   \\
                                            & Irregular            & IR     & 101   & 56       & 94   & 251   \\ \midrule
\multirow{3}{*}{DaG}    & Absent               & ABS    & 84    & 45       & 100  & 229   \\
                                            & Regular              & REG    & 156   & 60       & 118  & 334   \\
                                            & Irregular            & IR     & 173   & 98       & 177  & 448   \\ \midrule
\multirow{3}{*}{PIG}         & Absent               & ABS    & 253   & 112      & 223  & 588   \\
                                            & Regular              & REG    & 44    & 26       & 48   & 118   \\
                                            & Irregular            & IR     & 226   & 65       & 124  & 305   \\ \bottomrule
\end{tabular}}
\label{tab:statistics}
\end{table}

%% file: tables/table2.tex
\begin{table*}[ht!]
\caption{Comparisons against SOTA methods (test accuracy \%).  We only compare methods using the officially published Derm7t dataset train/validation/test split. Due to TFomer not specifying the number of experiment runs, for a fair comparison, we reproduce their results using the provided code and parameters, reporting the averages from 10 runs. The remaining results are all obtained from papers~\cite{bi2020multi, wang2022adversarial, tang2022fusionm4net}. The p-value (last column) is calculated from the AVG result by one-tailed paired t-tests between each method against SkinM2Former.}
\centering
\resizebox{\textwidth}{!}{
\begin{tabular}{@{}lllllllllll@{}}
\toprule
Method       & DIAG & BWV  & PN   & VS   & RS   & STR  & DaG  & PIG  & \textbf{AVG} & p-value\\
\midrule
Inception\_UB & 68.4 & 87.6 & 68.1 & 81.3 & 78.2 & 75.9 & 56.7 & 65.6 & 72.7 & 0.0045\\
Inception\_B  & 70.8 & 87.3 & 68.9 & 81.5 & 78.2 & 75.7 & 60.3 & 64.8 & 73.4 & 0.0023\\
Inception\_CB & 74.2 & 87.1 & 70.9 & 79.7 & 77.2 & 74.2 & 60.0 & 66.1 & 73.7 & 0.0002\\
MmA          & 70.6 & 83   & 65.6 & 75.7 & 73.9 & 69.4 & 59.2 & 61.3 & 69.8 & $8.5 \times 10^{-7}$\\
EmbeddingNet & 68.6 & 84.3 & 65.1 & 82.5 & 78.0 & 73.4 & 57.5 & 64.3 & 71.7 & 0.0011\\
TripleNet    & 68.6 & 87.9 & 63.3 & 83.0 & 76.0 & 74.4 & 61.3 & 67.3 & 72.7 & 0.0058\\
HcCNN        & 69.9 & 87.1 & 70.6 & \textbf{84.8} & 80.8 & 71.6 & 65.6 & 68.6 & 74.9 & 0.0310\\
AMFAM        & 75.4 & 88.1 & 70.6 & 83.3 & 80.8 & 74.7 & 63.8 & 70.9 & 76.0 & 0.0075\\
FusionM4Net  & 77.6$\pm$1.5 & 88.5$\pm$0.4 & 69.2$\pm$1.5 & 81.6$\pm$0.5 & \textbf{81.4$\pm$0.8} & 76.1$\pm$1.1 & 64.4$\pm$0.7 & 71.3$\pm$1.3 & 76.3$\pm$0.7 & 0.0490\\
TFormer      & 77.5$\pm$0.2 & 87.1$\pm$0.1 & 72.4$\pm$0.1 & 82.1$\pm$0.1 & 80.7$\pm$0.1 & 75.4$\pm$0.1 & 65.5$\pm$0.2 & 68.5$\pm$0.2 & 76.2$\pm$0.5 & 0.0096\\
SkinM2Former & \textbf{77.85±0.09}   & \textbf{88.83±0.09}          & \textbf{73.67±0.20} & 82.75±0.10          & 81.23±0.06 & \textbf{76.11±0.15} & \textbf{65.82±0.09} & \textbf{71.93±0.10} & \textbf{77.27±0.47} & NA \\
\bottomrule
\end{tabular}}
\label{tab:comparison}
\end{table*}

%% file: tables/table3.tex
\begin{table*}[ht!]
\centering
\footnotesize
\caption{Comparisons of top-performing SOTA methods (F1-score). The F1 score is calculated as the harmonic mean of the precision and recall scores. We only compare methods using the official Derm7t dataset with fixed train, validation, and test sets. TFormer results are replicated using their code and parameters, with averages from 10 experiments reported. The HcCNN did not give detailed results~\cite{bi2020multi}. The rest of the results are all obtained from papers~\cite{tang2022fusionm4net,yan2023towards,zhang2023tformer}. }
\begin{tabularx}{\textwidth}{@{}l *{9}{>{\centering\arraybackslash}X}@{}}
\toprule
Method       & DIAG & BWV  & PN   & VS   & RS   & STR  & DaG  & PIG  & \textbf{AVG}  \\ \midrule
Inception-UB & 43.92 & 76.39 & 63.88 & \textbf{61.23} & 64.96 & 60.15 & 53.77 & 48.04 & 59.04 \\
Inception-B  & 50.25 & 73.62 & 65.64 & 47.10 & 64.88 & 63.98 & 58.37 & 53.00 & 59.60 \\
Inception-CB & 63.40 & 80.63 & 68.31 & 50.62 & 72.40 & 63.33 & 58.60 & 56.26 & 64.19 \\
AMFAM        & 55.02 & 79.08 & 68.42 & 48.45 & 73.38 & 65.32 & 61.54 & 62.26 & 64.18 \\
FusionM4Net  & 57.70 & 80.58 & 68.67 & 42.88 & 72.26 & 67.35 & 63.34 & 56.64 & 63.68 \\
Tformer      & 66.86 & 78.88 & 71.15 & 57.04 & 73.87 & 66.00 & 63.09 & 61.82 & 67.34 \\
SkinM2Former & \textbf{66.88} & \textbf{81.22} & \textbf{72.15} & 56.57 & \textbf{74.41} & \textbf{67.38} & \textbf{63.62} & \textbf{64.85} & \textbf{68.38} \\ \bottomrule
\end{tabularx}
\label{tab:F1}
\end{table*}

%% file: sample_images/table_image1.tex
% \begin{table}[h!]
    \centering
    % First sub-table
    \resizebox{\textwidth}{!}{
        \setlength{\tabcolsep}{4pt} % Reduce space between columns
        \begin{tabular}{ccl}
            \hline
            \multirow{3}{*}{\textbf{cli}} & \multirow{3}{*}{\includegraphics[height=1.2cm,width=1.7cm,trim=1 1 1 1,clip]{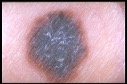}} & \multirow{6}{*}{\begin{tabular}[c]{@{}l@{}}
                \textbf{Diagnostic Difficulty}: high \\
                \textbf{Elevation}: palpable \\
                \textbf{Location}: upper limbs \\
                \textbf{Sex}: female \\
                \textbf{Management}: excision
            \end{tabular}} \\
            & & \\
            & & \\
            \multirow{3}{*}{\textbf{der}} & \multirow{3}{*}{\includegraphics[height=1.2cm,width=1.7cm,trim=1 1 1 1,clip]{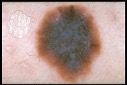}} & \\
            & & \\
            & & \\
            \hline
        \end{tabular}
    }

    % Second sub-table
    \resizebox{\textwidth}{!}{
        \setlength{\tabcolsep}{4pt} % Reduce space between columns
        \begin{tabular}{ccccc}
            % \hline
            \textbf{Label} & \textbf{GT} & \textbf{FusionM4Net} & \textbf{TFormer} & \textbf{SkinM2Former} \\
            \hline
            DIAG & BCC & \textcolor{red}{NEV} & BCC & BCC \\
            \hline
            PN & ABS & ABS & ABS & ABS \\
            BWV & ABS & ABS & ABS & ABS \\
            VS & ABS & ABS & \textcolor{red}{IR} & ABS \\
            PIG & ABS & ABS & \textcolor{red}{IR} & ABS \\
            STR & ABS & ABS & ABS & ABS \\
            DaG & IRG & IRG & IRG & IRG \\
            RS & ABS & ABS & ABS & ABS \\
            \hline
        \end{tabular}
    }
% \end{table}

%% file: sample_images/table_image2.tex
% \begin{table}[h!]
    \centering
    % First sub-table
    \resizebox{\textwidth}{!}{
        \setlength{\tabcolsep}{4pt} % Reduce space between columns
        \begin{tabular}{ccl}
            \hline
            \multirow{3}{*}{\textbf{cli}} & \multirow{3}{*}{\includegraphics[height=1.2cm,width=1.7cm,trim=1 1 1 1,clip]{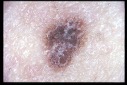}} & \multirow{6}{*}{\begin{tabular}[c]{@{}l@{}}
                \textbf{Diagnostic Difficulty}: medium \\
                \textbf{Elevation}: flat \\
                \textbf{Location}: buttocks \\
                \textbf{Sex}: male \\
                \textbf{Management}: excision
            \end{tabular}} \\
            & & \\
            & & \\
            \multirow{3}{*}{\textbf{der}} & \multirow{3}{*}{\includegraphics[height=1.2cm,width=1.7cm,trim=1 1 1 1,clip]{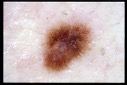}} & \\
            & & \\
            & & \\
            \hline
        \end{tabular}
    }

    % Second sub-table
    \resizebox{\textwidth}{!}{
        \setlength{\tabcolsep}{4pt} % Reduce space between columns
        \begin{tabular}{ccccc}
            \textbf{Label} & \textbf{GT} & \textbf{FusionM4Net} & \textbf{TFormer} & \textbf{SkinM2Former} \\
            \hline
            DIAG & NEV & NEV & NEV & NEV \\
            \hline
            PN & TYP & \textcolor{red}{ATP} & \textcolor{red}{ATP} & TYP \\
            BWV & ABS & ABS & ABS & ABS \\
            VS & ABS & ABS & ABS & ABS \\
            PIG & ABS & ABS & ABS & ABS \\
            STR & IR & \textcolor{red}{ABS} & \textcolor{red}{ABS} & IR \\
            DaG & IRG & \textcolor{red}{REG} & IRG & IRG \\
            RS & ABS & ABS & ABS & ABS \\
            \hline
        \end{tabular}
    }
% \end{table}

%% file: sample_images/table_image3.tex
% \begin{table}[h!]
    \centering
    % First sub-table
    \resizebox{\textwidth}{!}{
        \setlength{\tabcolsep}{4pt} % Reduce space between columns
        \begin{tabular}{ccl}
            \hline
            \multirow{3}{*}{\textbf{cli}} & \multirow{3}{*}{\includegraphics[height=1.2cm,width=1.7cm,trim=1 1 1 1,clip]{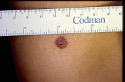}} & \multirow{6}{*}{\begin{tabular}[c]{@{}l@{}}
                \textbf{Diagnostic Difficulty}: medium \\
                \textbf{Elevation}: nodular \\
                \textbf{Location}: lower limbs \\
                \textbf{Sex}: female \\
                \textbf{Management}: excision
            \end{tabular}} \\
            & & \\
            & & \\
            \multirow{3}{*}{\textbf{der}} & \multirow{3}{*}{\includegraphics[height=1.2cm,width=1.7cm,trim=1 1 1 1,clip]{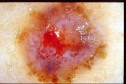}} & \\
            & & \\
            & & \\
            \hline
        \end{tabular}
    }

    % Second sub-table
    \resizebox{\textwidth}{!}{
        \setlength{\tabcolsep}{4pt} % Reduce space between columns
        \begin{tabular}{ccccc}
            \textbf{Label} & \textbf{GT} & \textbf{FusionM4Net} & \textbf{TFormer} & \textbf{SkinM2Former} \\
            \hline
            DIAG & MEL & MEL & MEL & MEL \\
            \hline
            PN & ABS & ABS & \textcolor{red}{ATP} & ABS \\
            BWV & PRS & ABS & ABS & PRS \\
            VS & IR & \textcolor{red}{ABS} & IR & IR \\
            PIG & IR & \textcolor{red}{ABS} & IR & IR \\
            STR & IR & \textcolor{red}{ABS} & IR & IR \\
            DaG & IRG & \textcolor{red}{ABS} & IRG & IRG \\
            RS & ABS & ABS & ABS & ABS \\
            \hline
        \end{tabular}
    }
% \end{table}

%% file: sample_images/table_image4.tex
% \begin{table}[h!]
    \centering
    % First sub-table
    \resizebox{\textwidth}{!}{
        \setlength{\tabcolsep}{4pt} % Reduce space between columns
        \begin{tabular}{ccl}
            \hline
            \multirow{3}{*}{\textbf{cli}} & \multirow{3}{*}{\includegraphics[height=1.2cm,width=1.7cm,trim=1 1 1 1,clip]{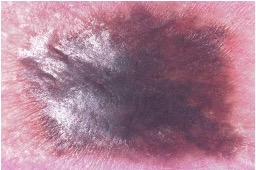}} & \multirow{6}{*}{\begin{tabular}[c]{@{}l@{}}
                \textbf{Diagnostic Difficulty}: high \\
                \textbf{Elevation}: palpable \\
                \textbf{Location}: chest \\
                \textbf{Sex}: female \\
                \textbf{Management}: excision
            \end{tabular}} \\
            & & \\
            & & \\
            \multirow{3}{*}{\textbf{der}} & \multirow{3}{*}{\includegraphics[height=1.2cm,width=1.7cm,trim=1 1 1 1,clip]{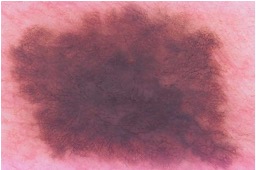}} & \\
            & & \\
            & & \\
            \hline
        \end{tabular}
    }

    % Second sub-table
    \resizebox{\textwidth}{!}{
        \setlength{\tabcolsep}{4pt} % Reduce space between columns
        \begin{tabular}{ccccc}
            \textbf{Label} & \textbf{GT} & \textbf{FusionM4Net} & \textbf{TFormer} & \textbf{SkinM2Former} \\
            \hline
            DIAG & MEL & MEL & MEL & MEL \\
            \hline
            PN & ATP & \textcolor{red}{ABS} & ATP & ATP \\
            BWV & ABS & ABS & ABS & ABS \\
            VS & ABS & ABS & ABS & ABS \\
            PIG & IR & IR & IR & IR \\
            STR & IR & IR & IR & IR \\
            DaG & REG & \textcolor{red}{IRG} & \textcolor{red}{IRG} & \textcolor{red}{IRG} \\
            RS & ABS & ABS & ABS & ABS \\
            \hline
        \end{tabular}
    }
% \end{table}

%% file: sample_images/table_image5.tex
% \begin{table}[h!]
    \centering
    % First sub-table
    \resizebox{\textwidth}{!}{
        \setlength{\tabcolsep}{4pt} % Reduce space between columns
        \begin{tabular}{ccl}
            \hline
            \multirow{3}{*}{\textbf{cli}} & \multirow{3}{*}{\includegraphics[height=1.2cm,width=1.7cm,trim=1 1 1 1,clip]{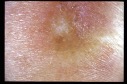}} & \multirow{6}{*}{\begin{tabular}[c]{@{}l@{}}
                \textbf{Diagnostic Difficulty}: medium \\
                \textbf{Elevation}: palpable \\
                \textbf{Location}: back \\
                \textbf{Sex}: male \\
                \textbf{Management}: clinical follow up
            \end{tabular}} \\
            & & \\
            & & \\
            \multirow{3}{*}{\textbf{der}} & \multirow{3}{*}{\includegraphics[height=1.2cm,width=1.7cm,trim=1 1 1 1,clip]{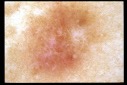}} & \\
            & & \\
            & & \\
            \hline
        \end{tabular}
    }

    % Second sub-table
    \resizebox{\textwidth}{!}{
        \setlength{\tabcolsep}{4pt} % Reduce space between columns
        \begin{tabular}{ccccc}
            \textbf{Label} & \textbf{GT} & \textbf{FusionM4Net} & \textbf{TFormer} & \textbf{SkinM2Former} \\
            \hline
            DIAG & MISC & \textcolor{red}{BCC} & MISC & MISC \\
            \hline
            PN & TYP & TYP & TYP & TYP \\
            BWV & ABS & ABS & ABS & ABS \\
            VS & ABS & ABS & ABS & ABS \\
            PIG & ABS & ABS & \textcolor{red}{IR} & ABS \\
            STR & ABS & ABS & ABS & ABS \\
            DaG & IRG & \textcolor{red}{REG} & IRG & IRG \\
            RS & ABS & ABS & \textcolor{red}{REG} & ABS \\
            \hline
        \end{tabular}
    }
% \end{table}

%% file: sample_images/table_image6.tex
% \begin{table}[h!]
    \centering
    % First sub-table
    \resizebox{\textwidth}{!}{
        \setlength{\tabcolsep}{4pt} % Reduce space between columns
        \begin{tabular}{ccl}
            \hline
            \multirow{3}{*}{\textbf{cli}} & \multirow{3}{*}{\includegraphics[height=1.2cm,width=1.7cm,trim=1 1 1 1,clip]{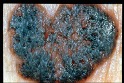}} & \multirow{6}{*}{\begin{tabular}[c]{@{}l@{}}
                \textbf{Diagnostic Difficulty}: low \\
                \textbf{Elevation}: palpable \\
                \textbf{Location}: back \\
                \textbf{Sex}: male \\
                \textbf{Management}: no further examination
            \end{tabular}} \\
            & & \\
            & & \\
            \multirow{3}{*}{\textbf{der}} & \multirow{3}{*}{\includegraphics[height=1.2cm,width=1.7cm,trim=1 1 1 1,clip]{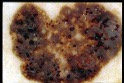}} & \\
            & & \\
            & & \\
            \hline
        \end{tabular}
    }

    % Second sub-table
    \resizebox{\textwidth}{!}{
        \setlength{\tabcolsep}{4pt} % Reduce space between columns
        \begin{tabular}{ccccc}
            \textbf{Label} & \textbf{GT} & \textbf{FusionM4Net} & \textbf{TFormer} & \textbf{SkinM2Former} \\
            \hline
            DIAG & SK & \textcolor{red}{MEL} & SK & SK \\
            \hline
            PN & ABS & ABS & \textcolor{red}{TYP} & ABS \\
            BWV & ABS & \textcolor{red}{PRS} & ABS & ABS \\
            VS & ABS & ABS & ABS & ABS \\
            PIG & IR & IR & IR & IR \\
            STR & ABS & ABS & ABS & ABS \\
            DaG & ABS & \textcolor{red}{IRG} & \textcolor{red}{IRG} & \textcolor{red}{IRG} \\
            RS & ABS & ABS & ABS & ABS \\
            \hline
        \end{tabular}
    }
% \end{table}

%% file: tables/table4.tex
\begin{table*}[t!]
\caption{The ablation study results of our SkinM2Former (test accuracy \%).}
\centering
\resizebox{\textwidth}{!}{
\begin{tabular}{@{}lllllllllll@{}}
\toprule
Method                & SW      & DIAG                & BWV                 & PN                  & VS                  & RS                  & STR                 & DAG                 & PIG                 & \textbf{AVG}           \\ \midrule
Swin+BCE (Baseline)             & \ding{108}           & 72.81±0.24          & 88.20±0.05          & 68.86±0.13          & 82.48±0.05          & 78.78±0.10          & 74.03±0.24          & 62.99±0.12          & 71.19±0.12          & 74.92±0.79          \\
TMCT+BCE              & \ding{108}          & 76.38±0.14          & 87.75±0.07          & 71.72±0.22          & 82.25±0.07          & 80.02±0.11          & 76.15±0.18          & 64.78±0.21          & 68.28±0.15          & 75.92±0.70          \\
TMCT+TWL              & \ding{108}          & 76.00±0.15          & 87.92±0.10          & 71.52±0.16          & 82.28±0.15          & 80.18±0.10          & 76.08±0.20          & 65.77±0.16          & 69.39±0.26          & 76.14±0.61          \\
TMCT+MHA+BCE          & \ding{108}          & 76.51±0.20          & 87.09±0.12          & 71.11±0.18          & 82.25±0.12          & 80.00±0.13          & 75.22±0.13          & \textbf{66.78±0.13} & 69.29±0.15          & 76.03±0.75          \\
TMCT(concat.image)+MHA+TWL   & \ding{108}   & 77.75±0.10   & \textbf{89.11±0.10} & 72.48±0.10          & 82.46±0.11          & 80.15±0.12          & 75.27±0.19          & 65.49±0.12          & 71.34±0.20          & 76.76±0.39 \\  
TMCT+MHA+TWL   & \ding{109}      & 77.22±0.11      & 88.08±0.10          & 72.18±0.13          & 82.41±0.10          & 79.70±0.17          & 75.04±0.14          & 64.71±0.16          & \textbf{72.23±0.15} & 76.44±0.68 \\ 
\textbf{TMCT+MHA+TWL(Ours)} & \ding{108} & \textbf{77.85±0.09} & 88.83±0.09    & \textbf{73.67±0.20} & 82.75±0.10          & \textbf{81.23±0.06} & \textbf{76.11±0.15} & 65.82±0.09          & 71.93±0.10 & \textbf{77.27±0.47} \\ 
\bottomrule
\end{tabular}
}
\label{tab:ablation}
\end{table*}

%% file: tables/table5.tex
\begin{table*}[t!]
\caption{Single vs. multi-modal skin lesion classification (test accuracy \%). }
\centering
\footnotesize
\resizebox{\textwidth}{!}{
\begin{tabular}{@{}llllllllll@{}}
\toprule
Modality     & DIAG                & BWV                 & PN                  & VS                  & RS                  & STR                 & DAG                 & PIG                 & \textbf{AVG}         \\ \midrule
Cli          & 62.23±0.06          & 84.81±0.10          & 57.87±0.20          & \textbf{84.81±0.10} & 74.78±0.14          & 65.32±0.15          & 54.70±0.18          & 61.01±0.14          & 67.62±0.69          \\
Der          & 71.75±0.16          & 88.21±0.06          & 66.58±0.14          & 81.57±0.07          & 79.59±0.12          & 75.80±0.10          & 60.76±0.04          & 70.08±0.17          & 74.29±0.39          \\
Meta         & 71.90±0.21          & 85.16±0.03          & 59.44±0.07          & 79.14±0.01          & 72.76±0.05          & 71.75±0.08          & 59.70±0.05          & 59.65±0.04          & 69.94±0.43          \\
Cli+Der      & 73.13±0.14          & 87.49±0.10          & 69.77±0.11          & 83.59±0.09          & 80.66±0.08          & 74.84±0.19          & 63.09±0.09          & 68.15±0.25          & 75.09±0.54          \\
Cli+Meta     & 71.80±0.16          & 85.67±0.05          & 63.80±0.11          & 80.66±0.17          & 75.54±0.17          & 69.82±0.18          & 60.10±0.16          & 61.32±0.32          & 71.09±0.60          \\
Der+Meta     & 77.32±0.06          & 88.56±0.09          & 73.06±0.18          & 81.32±0.09          & 79.44±0.07          & \textbf{76.71±0.26} & 65.67±0.19          & 71.65±0.08          & 76.72±0.42          \\
Cli+Der+Meta & \textbf{77.85±0.09} & \textbf{88.83±0.09} & \textbf{73.67±0.20} & 82.75±0.10          & \textbf{81.23±0.06} & 76.11±0.15          & \textbf{65.82±0.09} & \textbf{71.93±0.10} & \textbf{77.27±0.47} \\ \bottomrule
\end{tabular}}
\label{tab:single_multi}
\end{table*}

%% file: tables/table6.tex
\begin{table*}[t!]
\caption{Classification (test accuracy \%) using a final feature \( f_{final} \) obtained by concatenating different decision features ($f_{cli}'$, $f_{der}'$ and $f_{meta}'$). }
\centering
\footnotesize
\resizebox{\textwidth}{!}{
\begin{tabular}{@{}llllllllll@{}}
\toprule
Modality     & DIAG                & BWV                 & PN                  & VS                  & RS                  & STR                 & DAG                 & PIG            &     \textbf{AVG}        \\ \midrule
$f_{cli}'$          & 72.71±0.11          & 87.49±0.14          & 70.76±0.17          & 82.73±0.10          & 79.97±0.10          & 74.59±0.19          & 62.05±0.14          & 69.75±0.16          & 74.88±0.53          \\
$f_{der}'$          & 72.51±0.21          & 88.00±0.08          & 70.86±0.15          & 82.46±0.09          & 79.75±0.13          & 75.37±0.16          & 63.62±0.20          & 71.01±0.17          & 75.45±0.55          \\
$f_{meta}'$         & 76.38±0.09          & 88.84±0.04          & 66.91±0.02          & 80.51±0.11          & 77.75±0.06          & 72.78±0.13          & 63.49±0.13          & 68.89±0.16          & 74.43±0.49          \\
$f_{cli}'$+$f_{der}'$      & 72.05±0.11          & 87.92±0.08          & 70.03±0.11          & \textbf{83.39±0.10} & 79.44±0.17          & 74.68±0.12          & 62.46±0.14          & 69.37±0.21          & 74.92±0.64          \\
$f_{cli}'$+$f_{meta}'$     & 77.42±0.11          & \textbf{89.01±0.07} & 72.03±0.15          & 83.22±0.09          & 80.58±0.16          & 73.95±0.17          & 64.61±0.11          & 69.77±0.15          & 76.32±0.67          \\
$f_{der}'$+$f_{meta}'$     & 77.85±0.09          & 88.83±0.09          & \textbf{73.67±0.20} & 82.75±0.10          & \textbf{81.23±0.06} & \textbf{76.11±0.15} & \textbf{65.82±0.09} & \textbf{71.93±0.10} & \textbf{77.27±0.47} \\
$f_{cli}'$+$f_{der}'$+$f_{meta}'$ & \textbf{78.25±0.12} & 88.51±0.06          & 72.61±0.09          & 83.06±0.09          & 80.51±0.12          & 75.54±0.18          & 65.11±0.14          & 70.99±0.15          & 76.82±0.33          \\ \bottomrule
\end{tabular}
}
\label{tab:final_f}
\end{table*}